\pdfoutput=1

\documentclass[11pt]{article}

\usepackage[]{ACL2023}

\usepackage{times}
\usepackage{latexsym}

\usepackage{times}
\usepackage{amssymb}
\usepackage{bm}
\usepackage{arydshln}
\usepackage{latexsym}
\usepackage{bbm}
\usepackage{blindtext}
\usepackage{graphicx}
\usepackage{amsmath,amssymb}
\usepackage{booktabs}
\usepackage{multirow}
\usepackage[export]{adjustbox}

\usepackage{algorithm}
\usepackage{algpseudocode}

\usepackage[T1]{fontenc}

\usepackage[utf8]{inputenc}

\usepackage{microtype}

\usepackage{inconsolata}

\newcommand{\method}[1]{\textsc{#1}\xspace}
\newcommand{\dae}{\method{DAE}}
\newcommand{\bart}{\method{BART}}
\newcommand{\bartlargecnn}{\method{BART-large-cnn}}
\newcommand{\bartlargexsumsamsum}{\method{BART-large-samsum}}
\newcommand{\pegasus}{\method{PEGASUS}}
\newcommand{\pegasuscnn}{\method{PEGASUS-cnn}}
\newcommand{\pegasussamsum}{\method{PEGASUS-samsum}}
\newcommand{\condigsum}{\method{CONDIGSUM}}
\newcommand{\structurebart}{\method{S-BART}}
\newcommand{\gptthree}{\method{GPT-3}}
\newcommand{\weaksupclf}{\method{Weakly-Supervised-Classifier}}
\newcommand{\bertmulti}{\method{BertMulti}}
\newcommand{\encdecranker}{\method{EnDeRanker}}
\newcommand{\dialoglm}{\method{DialogLM}}
\newcommand{\qafacteval}{\method{QAFactEval}}
\newcommand{\tfive}{\method{T5}}
\newcommand{\tfivelargecnn}{\method{T5-large-cnn}}
\newcommand{\tfivelargesamsum}{\method{T5-large-samsum}}
\newcommand{\dataset}{\method{DiaSumFact}}

\title{Annotating and Detecting Fine-grained Factual Errors for Dialogue Summarization}

\author{Rongxin Zhu \quad Jianzhong Qi \quad Jey Han Lau \\
        School of Computing and Information Systems \\ 
        The University of Melbourne \\ 
        \texttt{rongxinz1@student.unimelb.edu.au, \{jianzhong.qi,laujh\}@unimelb.edu.au}}

\begin{document}
\maketitle
\begin{abstract}
A series of datasets and models have been proposed for summaries generated for well-formatted documents such as news articles. Dialogue summaries, however, have been under explored. 
In this paper, we present the first dataset with fine-grained factual error annotations named 
\dataset. 
We define fine-grained factual error detection as a sentence-level multi-label classification problem, and we evaluate two state-of-the-art (SOTA) models 
on our dataset.  Both models yield sub-optimal results, with a macro-averaged F1 score of around 0.25 over 6 error classes. We further propose an unsupervised model \textsc{\encdecranker} via candidate ranking using pretrained encoder-decoder models. Our model performs on par with the SOTA models while requiring fewer resources. These observations confirm the challenges in detecting factual errors from dialogue summaries, which call for further studies, for which our dataset and results offer a solid foundation.\footnote{The dataset and code are available at \url{https://github.com/731935354/Dia-Sum-Fact}}
\end{abstract}

\section{Introduction}
\begin{figure}[t]
\centering
\includegraphics[width=7cm]{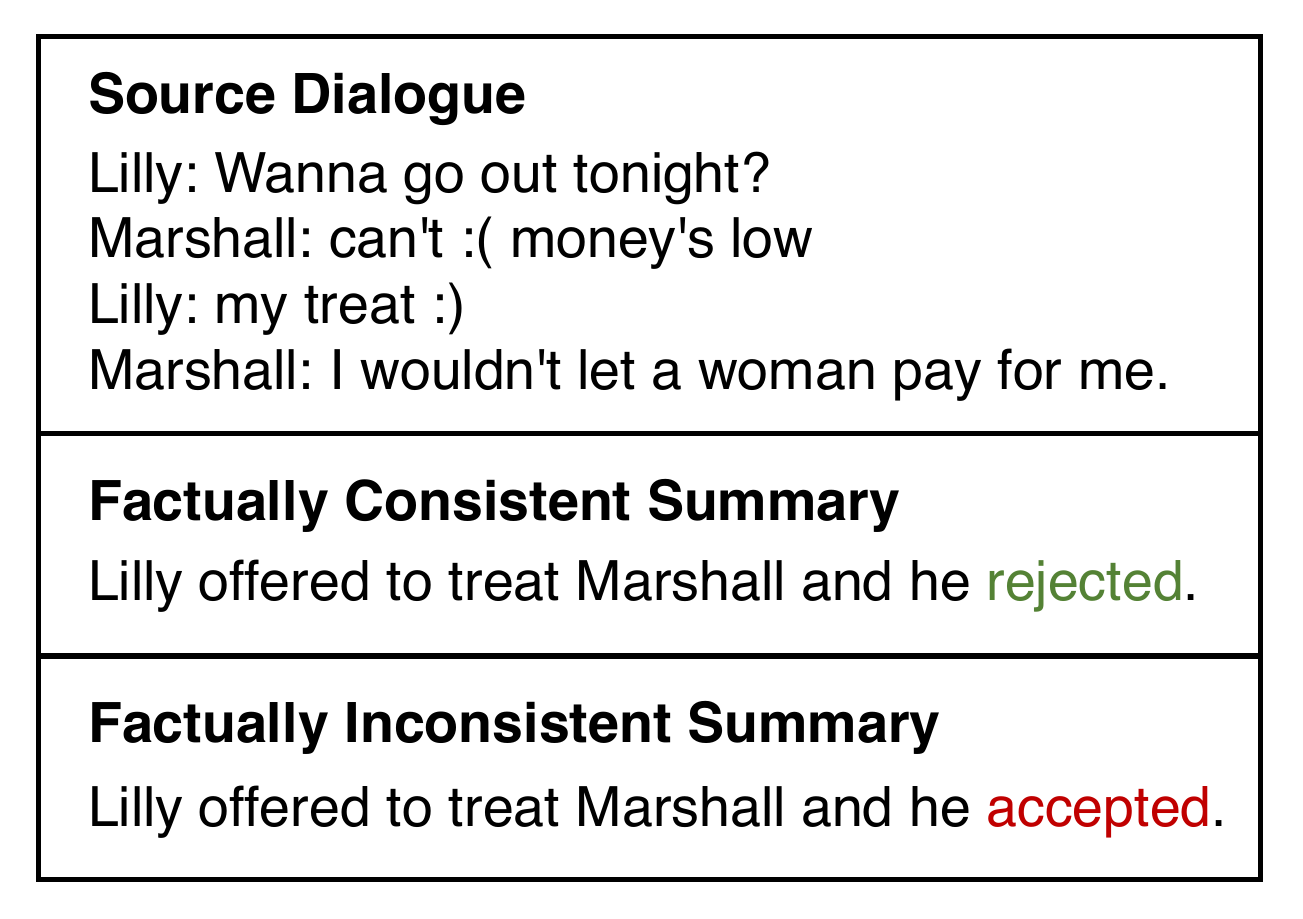}
\caption{Example summaries that are factually consistent and inconsistent with a source dialogue.}
\label{fig:example}
\end{figure}

Factual inconsistency in abstractive summarization --- a phenomenon where model-generated summaries contain facts that are inconsistent with the source document --- is a widely known problem and has been studied extensively in the document summarization community. An example is shown in Figure~\ref{fig:example}, where the source document is a dialogue --- the type of documents that this paper focuses on. 

Existing work covers topics on factual inconsistency  
including error typology and factuality annotations of state-of-the-art neural summarization models~\citep{maynez-etal-2020-faithfulness,huang-etal-2020-achieved,pagnoni-etal-2021-understanding,goyal-durrett-2021-annotating,fabbri-etal-2021-summeval,gao-wan-2022-dialsummeval,tang2022understanding}, automatic factual error detectors~\citep{wang-etal-2020-asking,goyal-durrett-2020-evaluating,kryscinski-etal-2020-evaluating,durmus-etal-2020-feqa,zeng-etal-2021-gradient,scialom-etal-2021-questeval}, methods to correct factual errors in summaries~\citep{cao-etal-2020-factual,dong-etal-2020-multi,chen2021improving} and methods to produce factually more consistent summaries~\citep{zhao-etal-2020-reducing,cao-wang-2021-cliff,tang-etal-2022-confit,zhu-etal-2021-enhancing,aralikatte-etal-2021-focus,chen-etal-2021-improving,balachandran2022correcting}. Almost all of these works focus on news summarization based on two datasets: \textsc{CNN/DailyMail}~\citep{hermann2015teaching,nallapati-etal-2016-abstractive2} and \textsc{XSum}~\citep{narayan-etal-2018-dont}.

Dialogue summarization (cf Figure~\ref{fig:example}), which aims to produce a condensed version of a dialogue while maintaining its salient information, is equally important due to its  application to summarizing meeting transcripts~\citep{li-etal-2019-keep,zhu-etal-2020-hierarchical,zhong2022dialoglm}, daily conversations~\citep{chen-yang-2020-multi,liu-chen-2021-controllable,feng-etal-2021-language}, customer service dialogues~\citep{liu2019automatic,zou2021unsupervised} and medical dialogues~\citep{joshi-etal-2020-dr,krishna-etal-2021-generating}. However,  factual consistency in dialogue summarization is under explored as there are currently no benchmark datasets that contain fine-grained error categories. This paper aims to fill in this gap. 

To investigate factual consistency in dialogue summarization, we release \dataset with fine-grained sentence-level annotations regarding factual consistency for 475 model summaries (1,340 sentences) from six neural dialogue summarization models on two popular datasets: \textsc{SAMSum}~\citep{gliwa-etal-2019-samsum} and \textsc{QMSum}~\citep{zhong-etal-2021-qmsum}. We adopt a two-dimensional typology that considers the semantic roles and verifiability of error spans separately. 

We formulate factual error detection as a sentence-level multi-label classification task and use \dataset to evaluate two state-of-the-art factual error detection models designed for document summarization. As there are no existing error detection model for fine-grained error categories, we adapt the two binary classification models to fit to our task. Empirical results show that they don't work well on the task, indicating its difficulty and the domain gap between document summarization and dialogue summarization.

We then propose two models: \bertmulti and \encdecranker. \bertmulti is a multi-class classification model trained on synthetic data, which is created by corrupting sentences from reference summaries~\citep{kryscinski-etal-2020-evaluating}. \encdecranker is a simple unsupervised model that can leverage any pretrained encoder-decoder model to detect factual errors. Given a model-generated summary sentence containing a span of interest for error detection, \encdecranker computes  log likelihood scores for the sentence and its variants containing replacement spans fetched from the source dialogue. The scores are computed as \textsc{BARTScore}~\citep{yuan2021bartscore}, which will be explained in ~\ref{sec:enderanker}. We compare the scores of the sentences to determine if the span of interest and hence the summary sentence contains a factual error. We run experiments with \tfive~\citep{raffel2020exploring}, \bart~\citep{lewis-etal-2020-bart} and \pegasus~\citep{zhang2020pegasus}, fine-tuned either on news summarization or dialogue summarization,  as the encoder-decoder for \encdecranker. The results show that \bertmulti and \encdecranker performs on par with the adapted state-of-the-art models in terms of macro-averaged F1.

Motivated by the strong complementarity between models, we further present two ensemble models combining the four models above. The results, while exceeding those of the individual models, are still far from indicating a practical model for factual error detection over dialogue summaries. This calls for further studies, for which our dataset and results form a solid foundation. 

To summarise, this paper makes the following contributions: 
\begin{itemize}
    \item We annotate and present \dataset, the first dataset with fine-grained sentence-level factual errors for dialogue summarization, providing rich annotation including error classes, erroneous spans and explanation.
    \item We investigate the effectiveness of adapting state-of-the-art factual error detection models for document summarization on model-generated dialogue summaries, demonstrating the difficulty of the task.
    \item We propose \bertmulti, a weakly-supervised multi-class classifier and \encdecranker, an unsupervised factual error detector that requires no human labeled data for training and can leverage existing pre-trained encoder-decoder models. Both models perform on par with adapted SOTA factual error detection models for document summarization.

 \item {Our experiments and analyses reveal the strengths and weaknesses of different factual error detection models, and point out future directions to improve them.}
\end{itemize}

\section{Related Work}
\textbf{Error typology and datasets.}  
There are a few existing datasets on factual errors. 
Some of them use binary (factually consistent or inconsistent) labels~\citep{kryscinski-etal-2020-evaluating, wang-etal-2020-asking} and 5-point Likert Scale labels~\citep{fabbri-etal-2021-summeval,gao-wan-2022-dialsummeval}, which require lower efforts to annotate, but they do not provide information on how and where factual errors were made. To support fine-grained analysis, multi-class and multi-dimensional typologies are designed. ~\citet{pagnoni-etal-2021-understanding} propose a linguistically motivated annotation framework that covers semantic frame errors, discourse errors and content verifiability errors. ~\citet{goyal-durrett-2021-annotating} use a 2-dimensional typology, where content verifiability and semantic error types are considered separately.  ~\citet{cao-etal-2022-hallucinated} focus on hallucinations and consider both factual and non-factual hallucination. ~\citet{tang2022understanding} unify different error types from previous works into a hierarchical taxonomy.
These datasets mostly focus on news summaries.

DialSummEval~\citep{gao-wan-2022-dialsummeval} is another popular dataset that contains annotation on factual consistency of model-generated dialogue summaries. The core difference of our work is that we consider fine-grained error categories and the text span (i.e., starting and ending position) of an error. Thus it provides a more elaborate, diagnostic assessment as to what and where goes wrong when a summary is not factually consistent. In comparison, DialSummEval only considers coarse-grained assessment of factuality using 5-point Likert Scale~\citep{joshi2015likert}, without specifying the actual error type (e.g., entity error).

\noindent\textbf{Factual error detection models.}
Most popular factual error detectors are based on either textual-entailment or question-answering (QA). 

Textual-entailment-based models are generally binary classifiers that take as input the source document and a model-generated summary. For example, ~\citet{kryscinski-etal-2020-evaluating} train binary factual error classifiers using synthetic data. ~\citet{zeng-etal-2021-gradient} use a gradient-based adversarial method to improve model accuracy. ~\citet{goyal-durrett-2020-evaluating} leverage dependency-level entailment achieving better performance and interpretability.

QA-based models first generate questions from a model-generated summary (or source dialogue), and then answer those questions based on its source dialogue (or a model-generated summary). The factual consistency is decided by the similarity between the ground truth answer and the predicted answer. For example, ~\citet{wang-etal-2020-asking,durmus-etal-2020-feqa} use a precision-oriented method that generates questions from model-generated summaries and answer them using the source document. ~\citet{scialom-etal-2019-answers} instead generate questions from a source document and answer them using the summary, making it a recall-oriented method. ~\citet{scialom-etal-2021-questeval} combine recall and precision-oriented techniques into a single framework. ~\citet{fabbri-etal-2022-qafacteval} refine the model component design and obtain a QA-based method that outperforms textual-entailment-based methods.

Our unsupervised method \encdecranker compares a span (e.g., a person name) in a model-generated sentence with candidates (e.g., other people's names in the dialogue) and decide the factual consistency of the span based on its rank among candidates. It achieves comparable macro F1 with adapted SOTA factual error detectors for document summarization but requires no labelled resources.

\section{The \dataset Dataset}
This section presents our \dataset dataset and procedures to construct the dataset. 

\subsection{Data Source}
To cover dialogues from different domains, we selected two popular datasets \textsc{SAMSum}~\citep{gliwa-etal-2019-samsum} and \textsc{QMSum}~\citep{zhong-etal-2021-qmsum}. \textsc{SAMSum} contains daily conversations and gold summaries. \textsc{QMSum} comes with queries and answers based on meeting transcripts. The answers to each query can be seen as a summary to an aspect of the meeting transcript.

For both \textsc{SAMSum} and \textsc{QMSum}, we randomly sampled 60 dialogues and their summaries in its test split.\footnote{For \textsc{QMSum} we also have the queries, in addition to the dialogues and summaries.}
For \textsc{QMSum}, we only chose queries whose gold utterances contain no more than 700 tokens according to Bert tokenizer.~\footnote{$50\%$ of the queries on aspects of meeting transcripts satisfy this constraint.} We manually filtered out dialogues with sensitive contents (e.g., dirty words and potential bias on gender or race). More statistics on the dataset can be found in Appendix Table~\ref{tab:dataset-stat} and Table~\ref{tab:semantic-err-freq}.

\begin{table*}[ht]
\centering
\begin{adjustbox}{max width=\linewidth}
\begin{tabular}{@{}llll@{}}
\toprule
Dialogue   & \multicolumn{2}{l}{\begin{tabular}[c]{@{}p{0.9\linewidth}@{}}Lucas: Where r u? I’m waiting at the airport.\\ Vanessa: There was a foul-up with the flight. I’m trying to get another ticket.\\ Lucas: OMG. How come?\\ Vanessa: No bloody idea. All of the flights are booked cos students are returning from holidays.\\ Lucas: I’ve called the airport and they said there’s a flight to New York at 9:45 p. m. \\ Vanessa: Great, I’ll book it now.\end{tabular}} \\ \midrule
\textbf{Error} & \textbf{Description}    & \textbf{Example Summary} & \textbf{In/Ex}   \\ \midrule
EntE       & \begin{tabular}[c]{@{}p{0.4\linewidth}@{}}The core arguments or their attributes in a semantic frame are wrong, such as the subjects and objects.\end{tabular} 
& \begin{tabular}[c]{@{}p{0.4\linewidth}@{}} \textit{\textcolor{red}{Vanessa} is waiting at the airport.} \end{tabular} & In  \\ \midrule
PredE      & \begin{tabular}[c]{@{}p{0.4\linewidth}@{}}The predicate, which is usually a verb, of a semantic frame is wrong.\end{tabular}   
& \begin{tabular}[c]{@{}p{0.5\linewidth}@{}}\textit{Lucas \textcolor{red}{has emailed} the airport and got some information about the flight to New York.}\end{tabular} & Ex \\ \midrule
CirE       & \begin{tabular}[c]{@{}p{0.4\linewidth}@{}}The non-core arguments, such as location modifiers, temporal modifiers are wrong.\end{tabular}                                   & \begin{tabular}[c]{@{}p{0.5\linewidth}@{}} \textit{Lucas is waiting at \textcolor{red}{the train station}.} \end{tabular} & Ex \\ \midrule
CorefE     & \begin{tabular}[c]{@{}p{0.4\linewidth}@{}}A pronoun or a reference (e.g., \texttt{this picture}) has a wrong antecedent or has no antecedents.\end{tabular}               & \begin{tabular}[c]{@{}p{0.5\linewidth}@{}}\textit{Vanessa is trying to get another ticket for \textcolor{red}{themselves}}.\end{tabular} & N/A  \\ \midrule
LinkE      & \begin{tabular}[c]{@{}p{0.4\linewidth}@{}}The relationship, e.g., a causal relationship, between  statements is wrong.\end{tabular}                                            & \begin{tabular}[c]{@{}p{0.5\linewidth}@{}}\textit{Vanessa will book the flight to New York at 9:45 pm \textcolor{red}{because students are returning from holidays}}.\end{tabular} & N/A \\ \midrule
Others       & \begin{tabular}[c]{@{}p{0.4\linewidth}@{}}This class covers the errors that do not fall into the above classes.\end{tabular}                                   & \begin{tabular}[c]{@{}p{0.5\linewidth}@{}} / \end{tabular} & N/A \\ \bottomrule
\end{tabular}
\end{adjustbox}
\caption{\label{table:error-typology} Factual error type descriptions and examples. \textbf{In/Ex} refers to Intrinsic Error (In) and Extrinsic Error (Ex).}
\end{table*}

\subsection{Summary Generation Models}
We generally choose models with publicly accessible pretrained model checkpoints or generated outputs instead of training models ourselves. 

On \textsc{SAMSum}, we use five models: \textbf{\bart}~\citep{lewis-etal-2020-bart}, \textbf{\pegasus}~\citep{zhang2020pegasus}, \textbf{\structurebart}~\cite{chen-yang-2021-structure}, \textbf{\condigsum}~\citep{liu-etal-2021-topic-aware} and \textbf{\gptthree}~\citep{brown2020language}. 
For  \textbf{\structurebart} and \textbf{\condigsum}, we obtain model outputs from the original papers.
For \textbf{\bart} and \textbf{\pegasus}, we generate 
output by running their pre-trained models.\footnote{We use \textit{linydub/bart-large-samsum} for \bart and \textit{transformersbook/pegasus-samsum} for \pegasus. Both are from \url{https://huggingface.co/models}.} For \textbf{\gptthree}, we fine-tune \textit{curie} over SAMSum dataset and generate summaries using the official API.\footnote{We fine-tuned it on May 27th, 2022 following \url{https://beta.openai.com/docs/guides/fine-tuning}.}

On \textsc{QMSum}, we use three models: \textbf{\pegasus}, \textbf{\bart} and \textbf{DialogLM}~\citep{zhong2022dialoglm}. Since we only focus on specific queries (i.e., queries that only ask about an aspect of a meeting, instead of summarizing the whole meeting), which is a subset of the original dataset, we fine-tuned them using specific queries only. The fine-tuned models achieve ROUGE scores that are better or comparable to state-of-the-art models on the complete dataset.\footnote{The ROUGE scores of the fine-tuned models are shown in Appendix A.2. We also tried 2-shot GPT-3 but found that it didn't work well in preliminary experiments and for that reason didn't include GPT-3.}

\subsection{Typology of Factual Errors}\label{sec:typology}
Motivated by~\citet{goyal-durrett-2021-annotating, pagnoni-etal-2021-understanding}, we adopt a 2-dimensional typology that treats semantic role and content verifiability of error spans separately.

On the semantic role dimension, we consider six error classes \textbf{Entity Error (EntE)}, \textbf{Predicate Error (PredE)}, \textbf{Circumstance Error (CirE)}, \textbf{Coreference Error (CorefE)}, \textbf{Link Error (LinkE)} and \textbf{Others}, with definitions and examples shown in Table~\ref{table:error-typology}. EntE, PredE, CirE are semantic frame errors, and CorefE, LinkE are discourse errors. When a sentence in the summary does not contain any factual error, we label it as \textbf{No Error}. 

For content verifiability, we consider \textbf{Intrinsic Error} (i.e., the error span consists of tokens from the source dialogue) and \textbf{Extrinsic Error} (i.e., the error span consists of tokens not mentioned in the source dialogue), a.k.a. hallucinations. This dimension is only defined for EntE, PredE and CirE.

\subsection{Annotation Procedure}\label{annotation-procedure}
We recruited 12 workers for the annotation task, including nine PhD students majored in natural language processing and three  Master's students majoring in linguistics and information technology. All annotators are fluent English speakers. We take an in-house annotation approach because a trial on Amazon Mechanical Turk did not yield meaningful results, even though high-quality crowd-sourced workers were sourced through strict constraints. The 12 annotators form six pairs randomly where each pair annotates 10 dialogues from each dataset. 

The annotation is done in three stages: pilot study, full annotation and annotation adjudication.

An annotation task involves analysing a dialogue and the summaries generated by all corresponding models. During the pilot study, annotators are required to go through the definition and examples for each error class to learn the labelling typology. Then, they will work on two pilot tasks, which are the same for all workers. For each task, a source dialogue and a model-generated summary are shown at the same time, and the annotator needs to label any factual errors in each individual sentence in the summary. When all sentences in the summary are done, another summary generated by a different model will be shown. Models are anonymized and their generations are shown in random order.

During the full annotation stage, we assign each annotator 10 tasks from each dataset, which are different from the tasks in pilot study. The annotations are only done for the semantic role dimension. 

In the adjudication stage, the two annotators of a pair along with an annotation quality controller (one of the authors of this paper) go through the annotations to resolve any disagreements,

and detailed notes were taken for reaching the final decisions (which is released as part of the dataset as it can  be useful for future analysis). Annotation mistakes are also corrected in this process. In the end, a total of 1340 sentences ($99.7\%$) with agreed annotations were obtained, while the rest of the sentences were discarded because no agreement can be made.

Note that the annotations on the content verifiability dimension are manually created by the annotation quality controller based on the detailed meeting notes of the last stage. It is a product of a post-annotation process because the original annotators did not explicitly label the error type as extrinsic or intrinsic. Instead, the annotators mark an \textbf{Extrinsic Error} for all error spans that are not mentioned in the source dialogue. The annotation quality controller takes this information and further split them into EntE, PredE and CirE based on the semantic role of an error span, and assign \textbf{Intrinsic Error} to all original EntE, PredE and CirE, thus obtaining a 2-dimensional annotation.

\subsection{Inter-annotator Agreement}
We use Cohen's Kappa~\cite{mchugh2012interrater} to evaluate the inter-annotator agreement. The scores in each group before adjudication are as follows. We first evaluate the agreement for binary label by merging all error types into a single negative class. The scores are 0.39, 0.44, 0.57, 0.59, 0.43, 0.51. For multi-class label, the scores are 0.34, 0.33, 0.44, 0.31, 0.31, 0.25. After adjudication we have full agreement for all instances (as explained in Section~\ref{annotation-procedure}).

\subsection{Results on the Summarization Models}
\begin{figure}[t]
\centering
\includegraphics[width=0.9\linewidth]{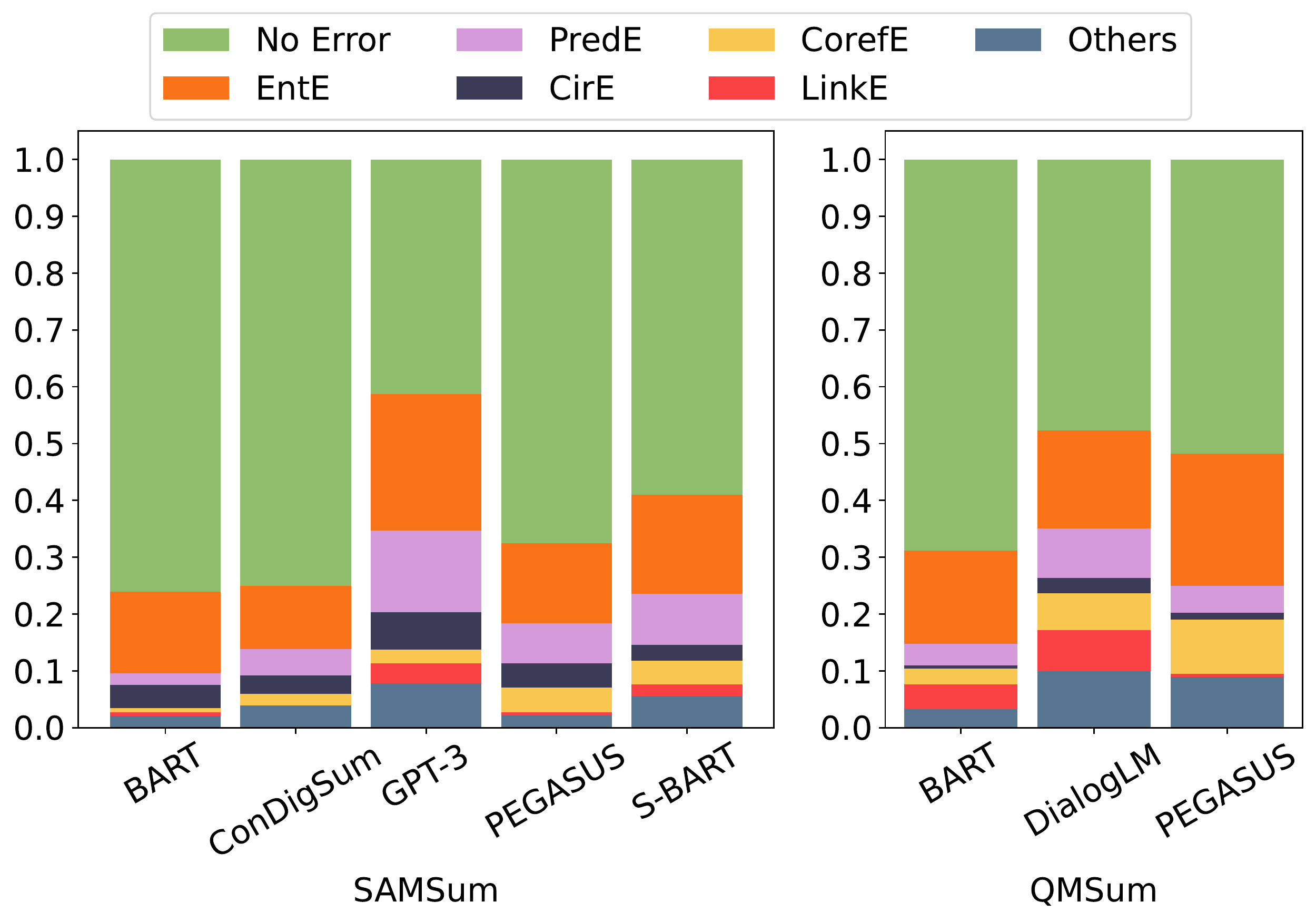}
\caption{Semantic factual error distribution of different summarization models on \textsc{SAMSum} and \textsc{QMSum}.}
\label{fig:semantic_frame_errors}
\end{figure}

\begin{figure}[t]
\centering
\includegraphics[width=0.9\linewidth]{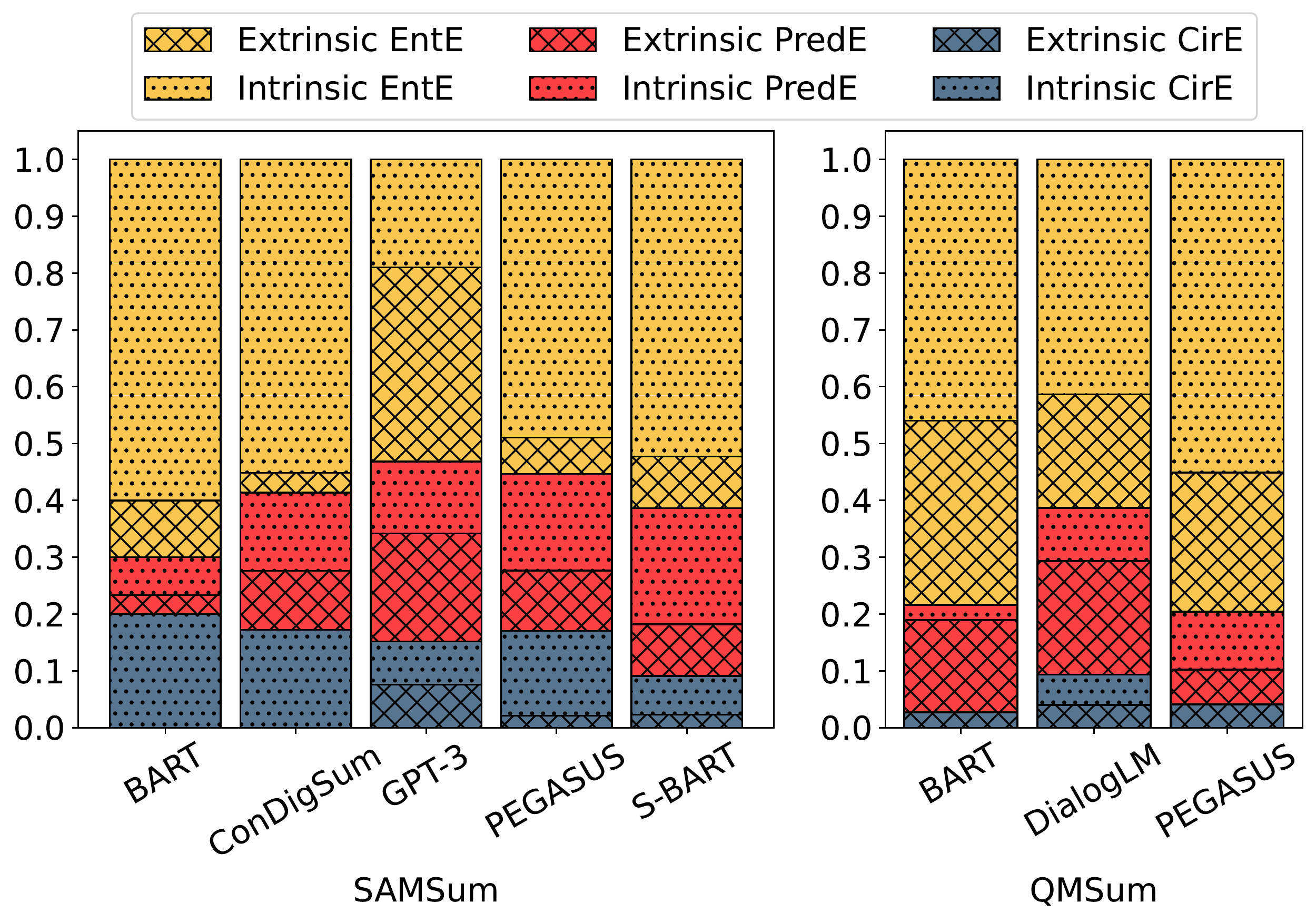}
\caption{Intrinsic and Extrinsic error distribution for EntE, PredE and CirE of different summarization models on \textsc{SAMSum} and \textsc{QMSum}.}
\label{fig:in-and-ex-errors}
\end{figure}

We summarize the performance results of the summarization models as derived from the annotations in this subsection. Figure~\ref{fig:semantic_frame_errors} and Figure~\ref{fig:in-and-ex-errors} show the factual error class distribution of the summarization models evaluated on \textsc{SAMSum} and \textsc{QMSum}. 

Overall, $33.3\%$ and $41.9\%$ sentences in model-generated summaries contain one or more factual errors in \textsc{SAMSum} and \textsc{QMSum}, respectively. The average number of errors for a factually inconsistent sentence is $1.14$. This indicates a broad existence of factual errors in the model-generated summaries, thus emphasizing the importance to resolve factual errors in dialogue summarization. 

Semantic frame errors (i.e., EntE, PredE and CirE) are more frequent than discourse errors (i.e., CorefE and LinkE) overall, while their distributions are not the same on both datasets.  
\textsc{SAMSam} has a higher portion of factually inconsistent sentences caused by semantic frame errors ($76.9\%$) than \textsc{QMSum} has ($58.9\%$), while \textsc{QMSum} has a higher portion of discourse errors ($24.0\%$) than \textsc{SAMSam} ($11.3\%$). We observe two main reasons for this discrepancy. First, the sentences in \textsc{QMSum} are longer and exhibit more complex discourse structures, especially causal relations, which can be challenging for models to summarize. Second, models fine-tuned on \textsc{QMSum} tend to copy large chunks of the input dialogue. Many pronouns are directly copied from the source dialogue without proper context, causing Coreference Errors (CorefE).

Among the different summarization models, \bart and \pegasus have been evaluated on both datasets where \bart generates summaries with fewer factual errors consistently. 
On \textsc{SAMSum}, $24.0\%$ of the sentences generated by \bart contain factual errors, which is the fewest, while the highest portion is reported by \gptthree, i.e., $58.7\%$. 
\condigsum and \structurebart are variants of \bart that achieve better ROUGE scores than \bart using contrastive learning and dialogue structure information, respectively. Our results reveal that both models produced more sentences with factual errors than \bart did, indicating that improvement in ROUGE may not help with the factual consistency of summaries. This result emphasizes the importance of more benchmark datasets for dialogue summarization model evaluation. On \textsc{QMSum}, \bart is still the best, while \dialoglm produced the highest proportion of sentences with factual errors.

On the content verifiability dimension, models on \textsc{QMSum} produce more extrinsic errors than on SAMSum. A potential reason is that reference summaries in \textsc{QMSum} contain more tokens outside the source dialogue. For \textsc{SAMSum}, all models are mainly dominated by intrinsic errors, while \gptthree produces more extrinsic errors than intrinsic ones.

\section{Detecting Factual Errors}

In this section, we automate factual error detection in model-generated summaries. We first adapt two state-of-the-art factual error detection models from document summarization. We then propose a weakly supervised multi-class classifier and a simple yet effective unsupervised model that can utilize any pretrained encoder-decoder model to identify factual errors. Finally, we present ensemble-based models combining all techniques above.  

\textbf{Problem statement.} We formulate factual error detection as a sentence-level multi-label classification task, i.e., given an input dialogue and a sentence from a model-generated summary, we classify whether the sentence contains any (semantic role) factual errors as outlined in Section~\ref{sec:typology}.

\subsection{Adapted State-of-the-Art Models}\label{sec:sota-models}

\textbf{\dae}~\citep{goyal-durrett-2020-evaluating} is based on dependency-level entailment, which predicts whether a dependency arc in a model-generated sentence is entailed by the input document (e.g., a dialogue in our problem). To adapt it to our problem, we design rules to map from dependency arc types to our factual error classes, as shown in Table~\ref{tab:dep-to-errcls}. Given a summary sentence, we use the trained \dae provided by the authors to predict dependency arcs in the sentence. The union of all factual error classes corresponding to the types of the predicted erroneous dependency arcs will be used as our factual error predictions. Note that not all factual error classes have corresponding dependency arc types and hence not all error classes can be detected by this model.

\begin{table}[t]
\setlength{\tabcolsep}{2pt}
\centering
\begin{tabular}{@{}lc@{}}
\toprule
Dependency Arc Types         & Error Class \\ \midrule
\begin{tabular}[c]{@{}p{0.70\linewidth}@{}}nsubj, obj, obl:agent, iobj, dobj, nmod, vocative, appos, nummod, compound, amod, det, clf, flat \end{tabular} & EntE \\ \midrule
obl:tmod, advmod & CirE    \\ \midrule
aux              & PredE   \\ \midrule
other arc types  & Others  \\ \bottomrule
\end{tabular}%
\caption{Rules to map from dependency arc types to our factual error classes.}
\label{tab:dep-to-errcls}
\end{table}

\begin{figure*}[ht]
\centering
\includegraphics[width=16cm]{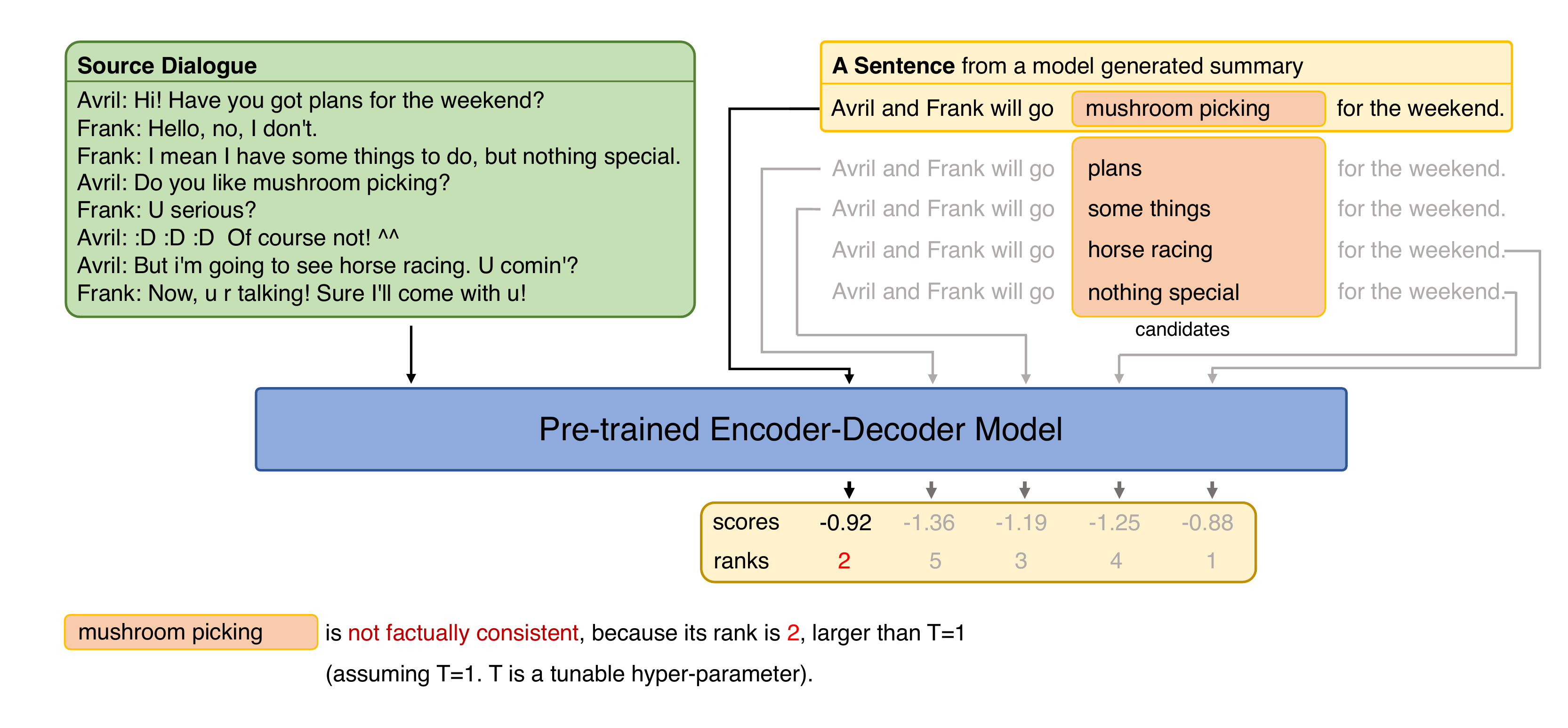}
\caption{The workflow of our \encdecranker model.}
\label{fig:encoder-decoder-ranker}
\end{figure*}

\textbf{QAFactEval}~\citep{fabbri-etal-2022-qafacteval} is a QA-based factual error detector. Given a question generation model (QG) and a question answering model (QA), which are trained on existing datasets for the question answering task, it works as follows: (1) Question-worthy spans ($s$), which are noun phrases and named entities, are extracted from a model-generated summary. (2) For each $s$, a question is generated by QG based on $s$ and the summary. (3) The QA model predicts an answer $a$ based on the question and the source document. (4) The similarity between $s$ and $a$ is measured by some metric. (5) The factual consistency of the summary is made based on the similarity scores for all $s$ in it.

We use the learned metric \texttt{LERC (QuIP)} mentioned in the paper and report a factual error if the similarity score between $s$ and $a$ is smaller than a threshold $T_{qa}$ (a hyper-parameter). Question-worthy spans of different semantic roles correspond to our semantic role-based factual error classes, as outlined in Algorithm~\ref{alg:sr2errcls} in Appendix. We obtain the semantic role of a question-worthy span by a pretrained structured prediction model in AllenNLP 2.9.3.\footnote{We use \textit{structured-prediction-srl-bert} and choose the semantic role of the shortest span containing $s$.}

\textbf{\weaksupclf} is a multi-class classifier that we construct. It takes as input a source dialogue and a generated summary sentence to predict factual error classes in the sentence, motivated by~\citet{kryscinski-etal-2020-evaluating}. We create synthetic training data by corrupting sentences in reference summaries as follows. 

For Entity Error, Circumstance Error and Coreference Error, we replace named entities or pronouns with those randomly picked from the same category. For Predicate Error, we replace verbs with other randomly chosen verbs. We match the form (e.g., tense) of the selected verbs to the original one. Negative replacements for all above classes are extracted from either the source dialogue or the whole dataset. For Link Error, we replace a discourse marker corresponding to causal relation (e.g., because) with another one indicating a reversed causal relation (e.g., so). More details on our synthetic data generation are in Appendix~\ref{sec:weak-supervised-classifier}.

We use cross entropy loss to train the classifier, which is based on BERT~\cite{devlin-etal-2019-bert} with a linear layer on top of \textsc{[CLS]} representation for classification. We concatenate the source dialogue and a sentence, delimited by \textsc{[SEP]}, as input. 

\begin{table*}[!ht]
\begin{adjustbox}{max width=\linewidth}
{%
\begin{tabular}{@{}lllllllll@{}}
\toprule
Model           & NoE        & EntE       & CirE       & PredE      & CorefE     & Others     & Micro Avg  & Macro Avg  \\ \midrule
\multicolumn{9}{c}{Adapted state-of-the-art models}                                                                                            \\ \midrule
\qafacteval        & $0.68_{0.04}$ & $\underline{0.45}_{0.03}$ & $\underline{0.23}_{0.11}$ & $0.00_{0.00}$ & $0.11_{0.06}$ & $0.00_{0.00}$ & $0.51_{0.03}$ & $0.25_{0.02}$ \\ 
\dae               & $0.77_{0.02}$ & $0.32_{0.05}$ & $0.03_{0.06}$ & $0.00_{0.00}$ & $0.00_{0.00}$ & $\underline{0.34}_{0.11}$ & $0.59_{0.02}$ & $0.24_{0.02}$ \\ \midrule\multicolumn{9}{c}{Weakly Supervised multi-class classifier}                                                                                \\ \midrule 
\bertmulti   & $0.72_{0.00}$ & $0.20_{0.00}$ & $0.08_{0.00}$ & $0.09_{0.00}$ & $\underline{0.29}_{0.00}$ & $0.08_{0.00}$ & $0.54_{0.00}$ & $0.24_{0.00}$ \\ \midrule
\multicolumn{9}{c}{\textbf{\encdecranker (ours)}}                                                                     \\ \midrule 
\bartlargecnn    & $0.67_{0.06}$ & $0.34_{0.07}$ & $0.04_{0.06}$ & $0.15_{0.04}$ & $0.12_{0.10}$ & $0.00_{0.00}$ & $0.47_{0.07}$ & $0.22_{0.01}$ \\
\bartlargexsumsamsum & $0.67_{0.06}$ & $0.35_{0.08}$ & $0.03_{0.04}$ & $0.21_{0.06}$ & $0.21_{0.13}$ & $0.00_{0.00}$ & $0.47_{0.05}$ & $0.24_{0.02}$ \\
\pegasuscnn      & $0.71_{0.03}$ & $0.37_{0.08}$ & $0.04_{0.05}$ & $0.18_{0.05}$ & $0.14_{0.09}$ & $0.00_{0.00}$ & $0.52_{0.04}$ & $0.24_{0.01}$ \\
\pegasussamsum    & $0.67_{0.04}$ & $0.37_{0.09}$ & $0.06_{0.07}$ & $0.19_{0.06}$ & $0.16_{0.11}$ & $0.01_{0.02}$ & $0.46_{0.05}$ & $0.24_{0.01}$ \\
\tfivelargecnn & $0.68_{0.04}$ & $0.35_{0.09}$ & $0.03_{0.04}$ & $0.15_{0.04}$ & $0.06_{0.03}$ & $0.01_{0.02}$ & $0.47_{0.05}$ & $0.21_{0.02}$ \\
\tfivelargesamsum   & $0.70_{0.08}$ & $0.35_{0.10}$ & $0.04_{0.05}$ & $\underline{0.22}_{0.08}$ & $0.14_{0.03}$ & $0.00_{0.00}$ & $0.51_{0.09}$ & $0.24_{0.03}$ \\ \midrule\multicolumn{9}{c}{Ensemble learning (\textbf{including our \encdecranker model})}                                                                                     \\ \midrule 
\textsc{FreqVoting}  & $0.79_{0.03}$ & $0.40_{0.05}$ & $0.05_{0.11}$ & $0.10_{0.08}$ & $0.12_{0.10}$ & $0.01_{0.02}$ & $\underline{0.62}_{0.03}$ & $0.24_{0.03}$ \\
\textsc{Logistic} & $\underline{0.80}_{0.03}$ & $0.44_{0.05}$ & $0.20_{0.13}$ & $0.00_{0.00}$ & $0.11_{0.10}$ & $0.03_{0.03}$ & $0.61_{0.03}$ & $\underline{0.26}_{0.04}$ \\ 
\bottomrule
\end{tabular}%
}
\end{adjustbox}
\caption{F1 scores for factual error detection models with a break down on each error class based on our annotated dataset \dataset. We report the average score and standard deviation over 5-fold cross validation. \textbf{Link Error (LinkE)} is merged into \textbf{Others} because almost no model can detect it. The best score for each column is \underline{underlined}.}
\label{tab:factual_error_detection_res_multiclass}
\end{table*}

\subsection{\encdecranker}\label{sec:enderanker}

Here, we present our proposed unsupervised model, \encdecranker.
Given a generated summary sentence, it first identifies a set of \textit{spans of interest} (SOI) which may correspond to factual errors. For each SOI, \encdecranker  replaces it with different candidate spans and calculates a score for each  span including the SOI.  The factuality of the SOI is then decided based on its score among the scores of all candidate spans. 
Figure~\ref{fig:encoder-decoder-ranker} summarizes the workflow of \encdecranker. Below we detail core steps of \encdecranker: (1) \textit{SOI identification}, (2)  \textit{candidate span generation}, (3) \textit{span scoring} and (4) \textit{ranking-based factual error detection}.

\textbf{Span of interest identification.}
An SOI is a snippet in a sentence for factual error classification. We consider noun phrases, named entities and verbs as SOIs, which are obtained using spaCy 3.1.4.\footnote{https://spacy.io/} We obtain the semantic roles of the SOIs like for \qafacteval, which will be used to decide the error class of an SOI later.

\textbf{Candidate span generation.}
For each SOI, we create a set of candidate spans  that can potentially replace it in the model generated summary sentence. For a named entity SOI, the candidate spans are entities of the same named entity class (e.g., \textbf{PERSON}) of the SOI extracted from the input dialogue. For the \textbf{PERSON} class, in particular, we include all speaker names on top of all other \textbf{PERSON} named entities extracted. 
For a verb SOI, we extract all verbs from the input dialogue according to the Part-of-Speech tags and match the form (e.g., tense) with the SOI. 
For a noun phrase SOI, all noun phrases from the input dialogue are considered as candidate spans. All candidate spans are extracted using spaCy 3.1.4.

\textbf{Span scoring.} 
Let $D$ be an input dialogue and $S$ be a generated summary sentence with $n$ tokens $\{w_1, w_2, \cdots, w_{n-1}, w_n\}$, which includes a candidate span or an SOI, denoted by $c$. We adopt a encoder-decoder model $\mathbb{M}$ to calculate a sentence score for $S$ conditioned on $D$ as follows, which is used as the score of span $c$, denoted by $\text{score}_{c}$. $\mathbb{M}$ can be any pre-trained encoder-decoder model, such as a summarization model.
\begin{equation}
\text{score}_c = \frac{1}{n}\sum_{i=1}^{n}\log{p(w_i|w_{< i}, D)}
\end{equation}
Intuitively, the score is the average log likelihood of each token $w_i$ in $S$, conditioning on the previous tokens in $S$ (i.e., $w_{<i}$) and $D$. Here, $w_{0}$ is the starting token of the decoder.

\textbf{Ranking-based factual error detection.} 
Given a set of candidate spans $C = \{c_1, c_2, \cdots, c_{|C|}\}$ of an SOI, we form $|C|$ sentences by replacing the SOI with each of the candidate spans. We calculate span scores for the SOI and the candidate spans, and rank the spans by their scores in descending order. If the SOI has a rank larger than a threshold $T$ (a hyper-parameter), we report it as erroneous and determine its error class based on its semantic role, as summarized in Algorithm~\ref{alg:sr2errcls} (cf. Appendix). The same process is repeated for all SOIs in $S$. The union of all error classes detected for the SOIs is the final factual error classes predicted for $S$.

\subsection{Ensemble Modeling}
We further build two simple ensemble models based on the four models above: Most \textbf{Freq}uent \textbf{Voting} (\textsc{FreqVoting}) and \textbf{Logistic regression} (\textsc{Logistic}). \textsc{FreqVoting} takes all predicted error classes from the four models above and uses the class(es) with the largest frequency as the final prediction. 
For \textsc{Logistic}, we train a logistic regression model for each factual error class that takes the binary outputs from the four models above as features.  We use the union of all factual error classes predicted by the different logistic regression models as the final prediction. 

\subsection{Experiments}
To evaluate the models described in the last section, we perform 5-fold cross validation~\citep{stone1978cross} using \dataset.\footnote{As it gives more reliable results considering the size of our dataset, compared to a usual train/test split.} Implementation details and parameter settings are discussed in Appendix~\ref{sec:error-detectors-details}.
We record the F1 scores (mean and standard deviation) of the models on each error class in Table~\ref{tab:factual_error_detection_res_multiclass}. 

\textbf{Results}: All models can detect EntE significantly and consistently better than the other classes. Different models show advantage on different error classes, while no model can outperform all the others on all error classes. 

\qafacteval performs the best on EntE (0.45) and CirE (0.23) but poorly on the other error classes. The reason is that only named entities and noun phrases are treated as question-worthy spans. Future work may consider question-worthy spans of different types, such as verbs and discourse markers, to cover more error classes. 

\dae performs well on EntE and Others, while it suffers on CirE, PredE and CorefE. The main reason is that not all error classes are covered in the rules mapping from dependency arc to error class. Since a dependency arc is related to two words, rule designing is not easy. Future work may leverage learned models to predict error class automatically.

\bertmulti shows the best results on CorefE ($0.29$) but poor performance on CirE, PredE and Others, despite its high performance on synthetic validation dataset ($0.98$  accuracy). It indicates the difference between synthetic and real factual errors.

Our proposed model \encdecranker  using different pretrained encoder-decoder models generally exhibits strong results on EntE, PredE and CorefE, while more improvements need to be done on CirE and Others. Among all variants of \encdecranker, \pegasuscnn performs on par with \qafacteval in terms of macro-averaged F1 score, while it does not require question generation and question answering models.

The two ensemble models improve on the micro and macro-averaged F1, indicating complementarity among the models. For most error classes, the ensemble models usually have the best or second best performance.

Overall, none of the models yielded a particularly high F1 score for any error class. It shows that fine-grained factual error detection in dialogue summaries is a challenging problem which calls for further studies, for which our results and dataset will serve as a solid foundation. 

\section{Conclusions}
We created a fine-grained multi-faceted dataset named \dataset on  factual consistency of dialogue summarization. \dataset offers insights into how and where current neural summarization models fail when they produce factually inconsistent details in dialogue summaries. It can also serve as a testbed for automating factual error detection. Our proposed error detection method, \encdecranker, is shown to perform on par with state-of-the-art models even though it requires no labelled training data. That said, we ultimately found that even ensembling several error detection methods do not produce results that are good enough for practical use, indicating opportunities for future research in this area.

\section{Limitations}
\encdecranker is only tested on \textsc{DiaSumFact}. Further tests on more datasets are required to establish its general applicability.

\section{Ethics Statement}
This study is conducted under the guidance of the ACL code of Ethics.
We manually filtered out potential offensive content and removed all information related to the identification of annotators. The annotators are all fairly paid based on the Australian minimum wage. The annotation protocol is approved under Human Ethics LNR Application with reference number 2022-24233-30104-3.

\section*{Acknowledgements}
This research was undertaken using the LIEF HPC-GPGPU Facility hosted at the University of Melbourne. This Facility was established with the assistance of LIEF Grant LE170100200. We want to thank Gisela Vallejo, Han Sun, Miao Li, Rui Xing, Wei Gao, Yanchuan Chang, Yulia Otmakhova, Zheng Wei Lim, Zhexi Li, Zhuohan Xie for their help in the annotation.

\bibliography{anthology,custom}
\bibliographystyle{acl_natbib}

\appendix

\section{Implementation Details}

\subsection{Cross Validation Settings}\label{sec:cross_val_setting}
We randomly split \dataset into 5 portions with equal number of examples and keep the splits consistent across all models. Each time we take one portion as the test set and combine the other four portions for training or validation, or both.  The details for the evaluation of each model are described below.
\begin{itemize}
    \item \bertmulti, \dae and \textsc{FreqVoting}: there is no hyper-parameter to tune. The model is only evaluated on different test sets for 5 times.
    \item \qafacteval and \encdecranker: they are unsupervised models so no training is needed. Each time the four portions are combined as validation set for hyper-parameter tuning. 
    \item \textsc{Logistic}: since this model requires supervised training, we combine the four portions, shuffle it and further split it into training set and validation set, following a ratio of 7:3. The validation set is used for hyper-parameter tuning.
\end{itemize}

\subsection{Summary Generation Models}
\textbf{\gptthree}: we use a batch size of 64 and fine-tune it for 2 epochs. During inference, \texttt{temperature} is set to $1.0$ and \texttt{max\_tokens} is set to $100$. The finetuned model achieves 41.7 and 15.9 on ROUGE-1 and ROUGE-2.

\noindent \textbf{\dialoglm}: we finetune \textit{MingZhong/DialogLED-large-5120} proposed in the original paper\footnote{\url{https://github.com/microsoft/DialogLM}}. We finetune it for 5 epochs using a batch size of 32 (per-device batch size is 2, gradient accumulation is 16) and learning rate $3\times 10^{-5}$. The fine-tuning takes 30 minutes. The finetuned model achieves $38.48$ and $13.70$ on ROUGE-1 and ROUGE-2, which are higher than $34.50$ and $9.92$ reported in the original paper.

\noindent \textbf{\pegasus}: we finetune \textit{google/pegasus-cnn\_dailymail} for 5 epochs using a batch size of 32 (per-device batch size is 2, gradient accumulation is 16) and learning rate $3\times 10^{-5}$. The fine-tuning takes 15 minutes. The finetuned model achieves $33.56$ and $11.35$ on ROUGE-1 and ROUGE-2.

\noindent \textbf{\bart}: we finetune \textit{facebook/bart-large-cnn} for 5 epochs using a batch size of 32 (per-device batch size is 2, gradient accumulation is 16) and learning rate $3\times 10^{-5}$. The fine-tuning takes 25 minutes. The finetuned model achieves $40.46$ and $14.93$ on ROUGE-1 and ROUGE-2.

All original models come from huggingface model hub\footnote{\url{https://huggingface.co/models}}. The fine-tuning for \bart, \pegasus and \dialoglm is conducted using \textit{run\_summarization.py} from Transformers~\footnote{\url{https://huggingface.co/docs/transformers/index}} 4.14.0.

During training, the input is the concatenation of the query and its relevant utterances, which is a subset of the whole meeting transcript. Utterances are concatenated as a long string, the query and utterances are delimited by ``||''.

\subsection{Error Detection Models}\label{sec:error-detectors-details}
\subsubsection{\weaksupclf}\label{sec:weak-supervised-classifier}
To obtain corrupted reference sentences with Entity Error, Coreference Error and Predicate Error, we first extract named entities, noun phrases and verbs using spaCy 3.1.4, then get their semantic roles like for \qafacteval in Section~\ref{sec:sota-models}. We finally map from semantic role to factual error class according to Algorithm~\ref{alg:sr2errcls}.

We generate 80k negative examples for each error class, among which 75k are used for training and 5k for validation. For EntE, PredE and CirE, the negative replacements for half of the data come from the same dialogue, while another half of the data uses negative replacements extracted from the whole dataset excluding the dialogue corresponding to the sentence. In this case we include both intrinsic and extrinsic negative replacements. Sentences from reference summaries are used for No Error. 

We use \textit{run\_glue.py} from Transformers 4.14.0 for model training. The pretrained model we use for \textsc{BERT} is \textit{bert-base-uncased}.

We tune batch size among 16, 32, 64 and 128. The best value is 64 according to the accuracy on validation set ($98.24\%$). The model is trained for 8 epochs and evaluated every 500 steps. The learning rate we use is $3\times10^{-5}$. The training takes 8 hours on a Tesla V100 GPU with 32GM RAM.
\subsubsection{\encdecranker}
The details of the pretrained models that we use are as follows: \\
\noindent \bartlargecnn: \textit{facebook/bart-large-cnn} \\
\noindent \bartlargexsumsamsum: \textit{lidiya/bart-large-xsum-samsum} \\
\noindent \pegasuscnn: \textit{google/pegasus-cnn\_dailymail} \\
\noindent \pegasussamsum: \textit{transformersbook/pegasus-samsum} \\
\noindent \tfivelargecnn: \textit{sysresearch101/t5-large-finetuned-xsum-cnn} \\
\noindent \tfivelargesamsum: We fine-tune it using \textit{run\_summarization.py} from Transformers 4.14.0 based on \textit{sysresearch101/t5-large-finetuned-xsum-cnn}. The final batch size is 2 with a gradient accumulation steps of 16 (i.e., the conceptual batch size is $2 \times 16 = 32$). The model is trained for 8 epochs on a single NVIDIA A100 (40G) GPU, taking 5 hours. We choose the batch size 32 among [8, 16, 32] because it produces the highest ROUGE-1 and ROUGE-2 on validation set.

\subsubsection{\dae}
We use the trained classifier provided by the authors of the \dae model\footnote{\url{https://github.com/tagoyal/dae-factuality}} and process each sentence in a model-generated summary separately. A dependency arc is considered as erroneous if the predicted probability for the positive class is less than~$0.5$.

\subsubsection{\qafacteval}
We use the model provided by the authors\footnote{\url{https://github.com/salesforce/QAFactEval}} and retrieve the similarity score between ground truth answers and predicted answers from logs, given by the learned model \texttt{LERC (QuIP)}. We tune the threshold $T_{qa}$ among [$0.5$, $1.0$, $1.5$, $2.0$] and choose $0.5$ as the final value, as it produces the highest macro-averaged F1 score.

The process to map from semantic role to factual error class is outlined in Algorithm~\ref{alg:sr2errcls}.

\begin{algorithm}[h]
\caption{Semantic Role to Factual Error Class. arg0 to arg5 are core semantic roles such as subject and object. `ARGM' is the prefix for non-core semantic roles such as ARGM-TMP (temporal modifier). V represents `verb'.}\label{alg:sr2errcls}
\begin{algorithmic}

\Require $s$  \Comment{a Span-of-Interest}
\Require $sr$ \Comment{the semantic role of $s$}
\State $pronouns \gets [\text{i}, \text{we}, \text{us}, \text{you}, \text{he}, \text{him}, \text{she}, \text{her},$ \\ $\text{it}, \text{they}, \text{them}, \text{this}, \text{that}, \text{these}, \text{those}, \text{myself},$ \\ $\text{yourself}, \text{himself}, \text{herself}, \text{ourselves}, \text{yourselves},$ \\ $\text{themselves}]$
\If{$sr$ in $[\text{arg0}, \text{arg1}, \text{arg2}, \text{arg3}, \text{arg4}, \text{arg5}]$}
    \If {$s \in pronouns$}
        \State Return CorefE
    \Else
        \State Return EntE
    \EndIf
\ElsIf{$sr$ contains `ARGM'}
    \State Return CirE
\ElsIf{$sr = \text{`V'}$}
    \State Return PredE
\Else
    \State Return Others
\EndIf
\end{algorithmic}
\end{algorithm}

\subsubsection{\encdecranker}
We tune T (i.e., the rank threshold) among [1, 2, 3, 4, 5, 6, 7, 8, 9, 10] and choose the smallest value that achieves the highest macro-averaged F1 on the validation set. The best T values for different pre-trained models are as follows:
\begin{itemize}
    \item \bartlargecnn: 2
    \item \bartlargexsumsamsum: 2
    \item \pegasuscnn: 3
    \item \pegasussamsum: 2
    \item \tfivelargecnn: 3
    \item \tfivelargesamsum: 3
\end{itemize}
To avoid repeated encoding for the same dialogue, which corresponds to multiple sentences for factual error detection, we cache the encoded representation in encoder and reuse them to improve inference speed.

The experiments are conducted on a single Nvidia V100 GPU with 16GM RAM. The inference over a full pass of our dataset takes around 40 hours with a batch size of 1. The computaional overhead can be reduced by (1) reducing the number of Span of Interest (SOI) in a sentence, and (2) reducing the number of candidates, especially for noun phrases. We also tried distilled encoder-decoder models, but the results are sub-optimal.

\subsubsection{Ensemble Learning}
For ensemble models (i.e., \textsc{FreqVoting} and \textsc{Logistic}), the best \encdecranker is chosen based on the performance on the validation set, which is $30\%$ of the four portions combined except the test set, as introduced in~\ref{sec:cross_val_setting}.

During the training of \textsc{Logistic}, we upsample the minority class to match the number of the majority class for each logistic regression model corresponding to different factual error types.

\begin{table*}[ht]
\centering
\begin{tabular}{@{}lrrrll@{}}
\toprule
Dataset                                                             & \#Mod & \#Summ & \#Sen & Domain   & Annotation Typology               \\ \midrule
\begin{tabular}[c]{@{}l@{}}FactCC\\~\citep{kryscinski-etal-2020-evaluating}\end{tabular} & 10 & /    & 1,434 & news     & binary (consistent, inconsistent) \\
\begin{tabular}[c]{@{}l@{}}QAGS\\~\citep{wang-etal-2020-asking}\end{tabular}            & 2        & 474         & /           & news     & binary (consistent, inconsistent) \\
\begin{tabular}[c]{@{}l@{}}SummEval\\~\citep{fabbri-etal-2021-summeval} \end{tabular}   & 44       & 12,800       & /           & news     & 5-point Likert scale              \\
\begin{tabular}[c]{@{}l@{}}Polytope\\~\citep{huang-etal-2020-achieved}\end{tabular}     & 10       & 1,500        & /           & news     & multi-class                       \\
\begin{tabular}[c]{@{}l@{}}Cao'22\\~\citep{cao-etal-2022-hallucinated}\end{tabular}         & 1        & 800         & /           & news     & multi-class                       \\
\begin{tabular}[c]{@{}l@{}}Maynez'20\\~\citep{maynez-etal-2020-faithfulness}\end{tabular}       & 5        & 500         & /           & news     & binary (intrinsic, extrinsic)     \\
\begin{tabular}[c]{@{}l@{}}Frank\\~\citep{pagnoni-etal-2021-understanding}\end{tabular} & 8        & 2,250        & 4,942        & news     & multi-class                       \\
\begin{tabular}[c]{@{}l@{}}Goyal'21\\~\citep{goyal-durrett-2021-annotating}\end{tabular}       & 3        & 50          & /           & news     & multi-dimensional, multi-class    \\
\begin{tabular}[c]{@{}l@{}}CLIFF\\~\citep{cao-wang-2021-cliff}\end{tabular}            & 2        & 600         & /           & news     & multi-class                       \\
\begin{tabular}[c]{@{}l@{}}ConFIT\\~\citep{tang-etal-2022-confit}\end{tabular}       & 4        & 76          & /           & dialogue & multi-class                       \\
\begin{tabular}[c]{@{}l@{}}DialSummEval\\~\citep{gao-wan-2022-dialsummeval}\end{tabular} & 13 & 4,200 &      & dialogue & 5-point Likert Scale              \\
\textbf{\textsc{DiaSumFact} (ours)}                                                & 6        & 475         & 1,340        & dialogue & multi-dimensional, multi-class    \\ \bottomrule
\end{tabular}%
\caption{Datasets that focus on or include factual consistency for summarization. \#Mod: the number of summarization models covered. \#Summ: the number of model-generated summaries covered. \#Sen: the total number of sentences in model-generated summaries.}
\label{tab:my-table}
\end{table*}

\section{Data Annotation}
\subsection{Error Typology}
For CorefE, if a reference comes without antecedents in the input dialogue, we ignore the error in the summary.

\subsection{Annotation Tool}
\begin{figure*}[ht]
\centering
\includegraphics[width=\linewidth, frame]{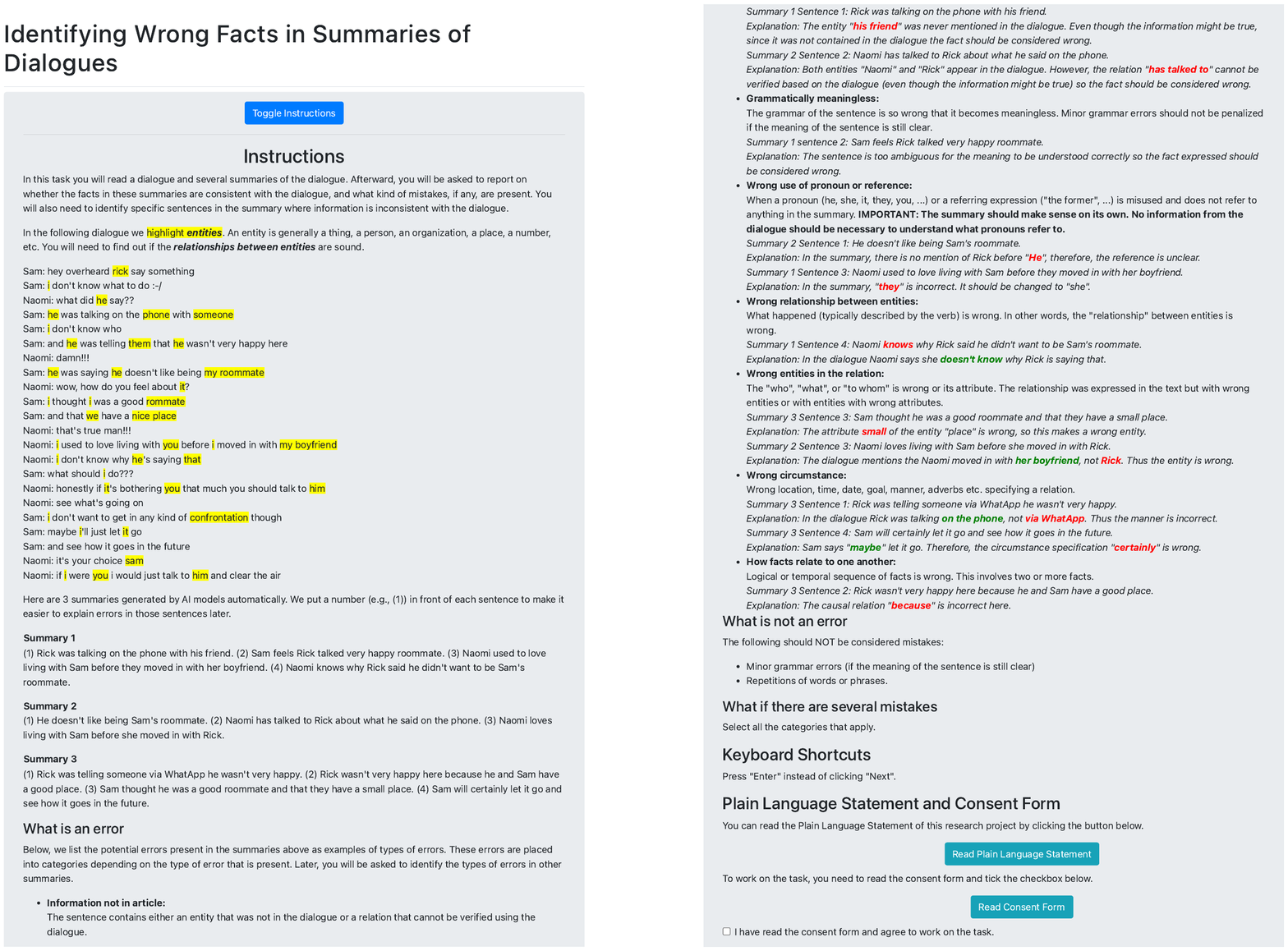}
\caption{The instruction page of our annotation tool, where an example dialogue, definitions and examples of different types of factual errors are shown. The plain language statement and consent form are at the bottom.}
\label{fig:anno_interface_instruction}
\end{figure*}

\begin{figure*}[ht]
\centering
\includegraphics[width=\linewidth, frame]{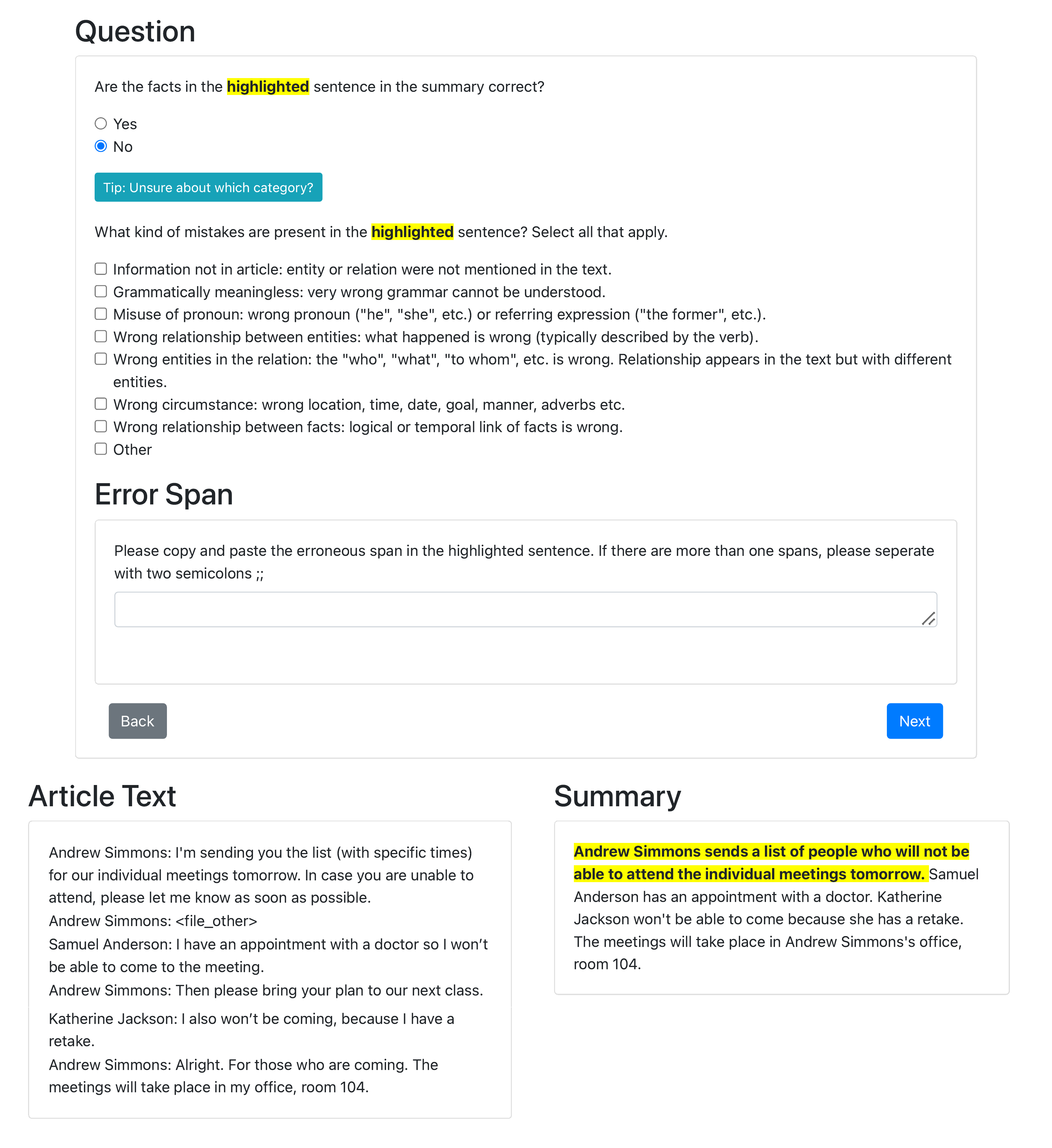}
\caption{The main page of our annotation tool.}
\label{fig:anno_interface_main_page}
\end{figure*}

\begin{figure*}[ht]
\centering
\includegraphics[width=\linewidth, frame]{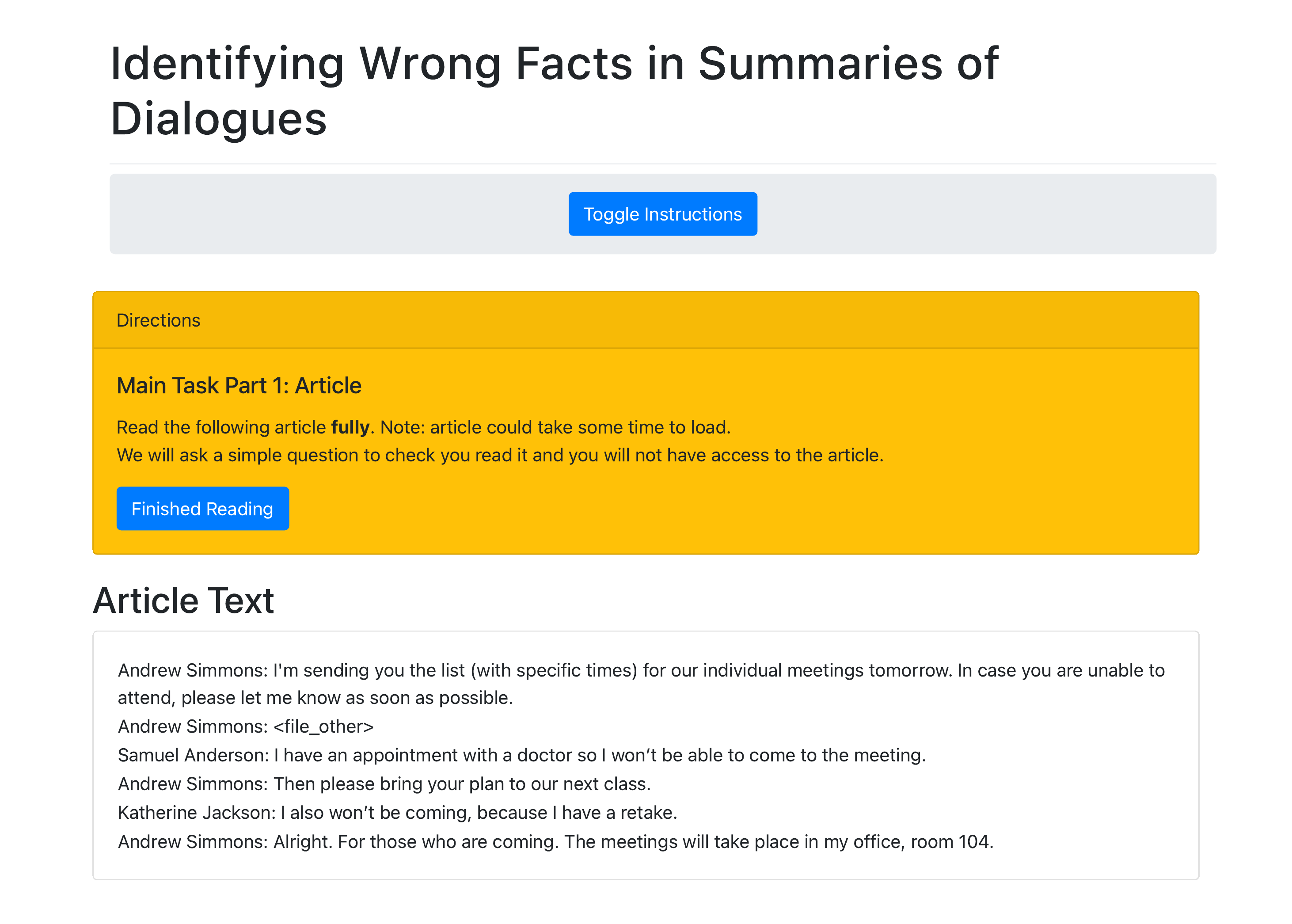}
\caption{The entity question page (part 1) of our annotation tool. Annotators are required to answer the entity question first to make sure they read the dialogue carefully.}
\label{fig:anno_interface_entity_question_part1}
\end{figure*}

\begin{figure*}[ht]
\centering
\includegraphics[width=\linewidth, frame]{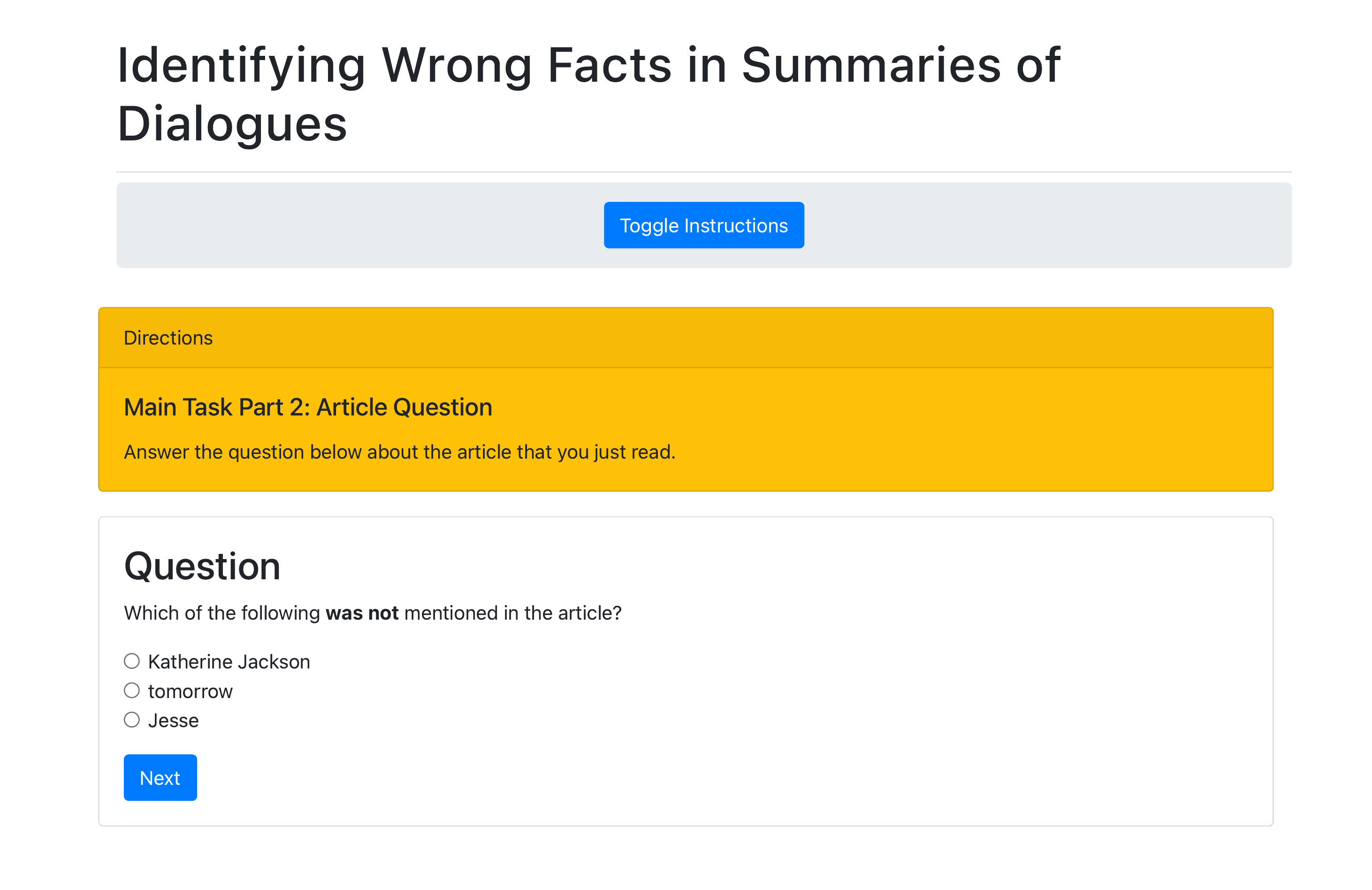}
\caption{The entity question page (part 2) of our annotation tool. Annotators are required to answer the entity question first to make sure they read the dialogue carefully.}
\label{fig:anno_interface_entity_question_part2}
\end{figure*}

We modify a web application developed originally for FRANK~\citep{pagnoni-etal-2021-understanding}\footnote{\url{https://github.com/artidoro/frank-annotation-platform}} to fit to our task. Specifically, we replace the example article and model summaries with an example dialogue and manually composed summaries to help explain different error types. We also add an input field for error span annotation in the main page. Screenshots are shown in Figures~\ref{fig:anno_interface_instruction}, \ref{fig:anno_interface_main_page}, \ref{fig:anno_interface_entity_question_part1} and \ref{fig:anno_interface_entity_question_part2}.

For in-house annotation, we deploy the web application on Firebase\footnote{\url{https://firebase.google.com/}} and provide with annotators URLs to the tasks directly.

\subsection{Annotation Procedure}
The initial annotation by all annotators follows the typology proposed by \citet{pagnoni-etal-2021-understanding}, which includes two additional classes: Out-of-Article Error (i.e., Extrinsic Error in our paper) and Grammar Error. We merge Grammar Error to Others, and treat Extrinsic Error as a separate dimension, as outlined in~\ref{sec:typology}.

\subsection{Payments to Annotators}
All our annotators are volunteers. We pay 100 AUD to each annotator. The annotation task begins after they agree to the amount of payment.

\subsection{Demographic Characteristics of Annotators}
1 annotator come from Colombia, 1 annotator comes from Russia, 1 annotator come from Malaysia, 9 annotators come from China. There are 6 female and 6 male annotators.

\subsection{Consent from Annotators}
We show the consent form in the annotation web application. Annotation can only begin after consent form is received from annotators.

\section{Case Study}
\begin{figure*}[ht]
\centering
\includegraphics[width=\linewidth, frame]{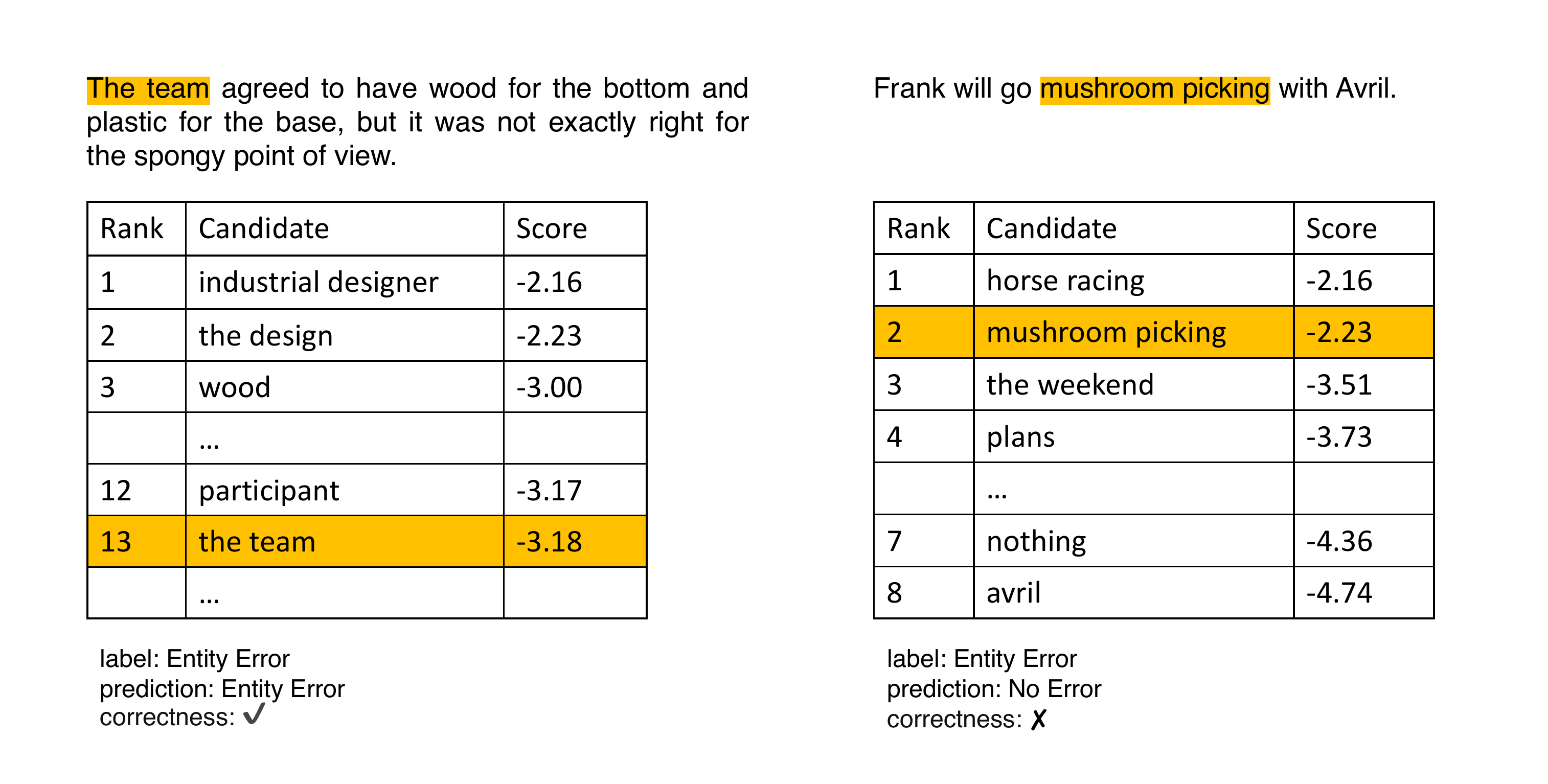}
\caption{Case study for \textsc{EnDeRanker} where it identifies an error correctly in the example on the left, but fails in the right-side example. The rank threshold T=3. The SOIs are highlighted both in the original sentence and in the candidates list sorted by score in descending order.}
\label{fig:case-enderanker}
\end{figure*}

As shown in Figure~\ref{fig:case-enderanker}, our \encdecranker successfully identifies an error of the span ``The team'' because its rank is larger than the threshold $T=3$. Since the semantic role of the span is \textit{arg0}, the model predicts Entity Error according to Algorithm~\ref{alg:sr2errcls}. On the right-side example, \encdecranker fails to report the error of ``muchroom picking'', although the factual consistent span ``horse racing'' is ranked at the top among candidates. The reason is that $T$ is too large. For future work, we may design error identification methods using SOI-specific thresholds rather than a universal threshold for all SOIs.

\section{Potential Risks}
The factual error detection models we propose, which are \bertmulti and \encdecranker, do not produce satisfactory performance to be used for real applications. We do not advise people to use them directly in real applications as factual error detectors for dialogue summarization without further improvements.

\section{Intended Use of Existing Artifacts}
The SAMSum~\citep{gliwa-etal-2019-samsum} dataset is shared on terms of the AttributionNonCommercial-NoDerivatives 4.0 International (CC BY-NC-ND 4.0) license. We provide additional information (i.e., model-generated summaries and human annotations) without modifying the original data (i.e., dialogues and reference summaries).

\begin{table}[ht]
\centering
\begin{adjustbox}{max width=\linewidth}
\begin{tabular}{@{}lll@{}}
\toprule
Data source & \textsc{SAMSum} & \textsc{QMSum} \\ \midrule
\#Exs       & 757                              & 583                             \\
$T_{D}$     & 148.4                            & 355.7                           \\
$U_{D}$     & 12.3                             & 16.2                            \\
$T_{Sen}$   & 11.3                             & 25.7                            \\
$S_{Summ}$  & 2.6                              & 3.2                             \\
$T_{Q}$     & /                                & 14.9                            \\ \bottomrule
\end{tabular}%
\end{adjustbox}
\caption{Statistics of our dataset. \#Exs: the number of (dialogue, sentence) pairs. $T_{D}$: the average number of tokens in a dialogue. $U_{D}$: the average number of utterances in a dialogue. $T_{sen}$: the average number of tokens in a summary sentence. $S_{Summ}$: the average number of sentences in a model-generated summary. $T_{Q}$: the average number of tokens in a query.}
\label{tab:dataset-stat}
\end{table}

\begin{table}[]
\centering
\begin{tabular}{@{}ll@{}}
\toprule
Error Type & Frequency \\ \midrule
No Error   & 853       \\ 
EntE       & 256       \\ 
PredE      & 106       \\ 
CirE       & 48        \\ 
CorefE     & 62        \\ 
LinkE      & 41        \\ 
Others     & 42        \\ \bottomrule
\end{tabular}%
\caption{The number of human-detected errors in each error type along semantic role dimension.}
\label{tab:semantic-err-freq}
\end{table}

\end{document}